\RequirePackage{fix-cm}
\documentclass[twocolumn]{svjour3} 

\smartqed

\usepackage{subcaption}
\usepackage{times}
\usepackage{epsfig}
\usepackage{graphicx}
\usepackage{amsmath}
\usepackage{amssymb}
\usepackage{balance}
\usepackage{color}
\usepackage{siunitx}
\usepackage{multirow}
\usepackage{stfloats}
\usepackage{comment}
\usepackage{listings}
\usepackage{algorithm}
\usepackage{wrapfig}
\usepackage{epsfig}
\usepackage[utf8]{inputenc} % allow utf-8 input
\usepackage[T1]{fontenc}    % use 8-bit T1 fonts
\usepackage{siunitx}
\usepackage{cite}
\usepackage{ctable}
\usepackage{hhline}
\usepackage{booktabs}
\usepackage{pifont}
\usepackage{csquotes}

\usepackage[round, sort]{natbib}
\usepackage[pagebackref=true,breaklinks=true,letterpaper=true,colorlinks,bookmarks=false]{hyperref}
\usepackage{color, colortbl}
\definecolor{Gray}{gray}{0.9}
\definecolor{Red}{RGB}{230, 57, 70}

\newlength\savewidth

\begin{document}
\sloppy
% \begin{flushend}
\title{KptLLM++: Towards Generic Keypoint Comprehension with Large Language Model}

\author{Jie Yang \and
        Wang Zeng \and
        Sheng Jin \and
        Lumin Xu \and
        Wentao Liu$^\dag$ \and
        Chen Qian \and \\ Zhen Li \and Ruimao Zhang$^\dag$}
        
\institute{ Jie Yang is with Sun Yat-sen University, Shenzhen, and also affiliated with the Chinese University of Hong Kong, Shenzhen. \\ 
              \email{jieyang5@link.cuhk.edu.cn}\\
           \and
                Wang Zeng, Sheng Jin, Wentao Liu, and Chen Qian are with SenseTime Research and Tetras.AI. \\
              \email{zengwang, jinsheng, liuwentao, qianchen@tetras.ai}\\
           \and           
           Lumin Xu is with the Chinese University of Hong Kong. \\
              \email{luminxu@link.cuhk.edu.hk}\\
           \and
           Zhen Li is with the Chinese University of Hong Kong, Shenzhen. \\
              \email{lizhen@cuhk.edu.cn}\\
           \and
           Ruimao Zhang is with Sun Yat-sen University, Shenzhen, and also with the Guangdong Key Laboratory of Big Data Analysis and Processing, Guangzhou.\\
              \email{ruimao.zhang@ieee.org} 
              $^\dag$ denotes the corresponding author.
}

\date{Received: date / Accepted: date}

\maketitle

\begin{abstract}

The emergence of Multimodal Large Language Models (MLLMs) has revolutionized image understanding by bridging textual and visual modalities. However, these models often struggle with capturing fine-grained semantic information, such as the precise identification and analysis of object keypoints. Keypoints, as structure-aware, pixel-level, and compact representations of objects, particularly articulated ones, play a crucial role in applications such as fine-grained image analysis, object retrieval, and behavior recognition.
In this paper, we propose KptLLM++, a novel multimodal large language model that specifically designed for generic keypoint comprehension through the integration of diverse input modalities guided by user-defined instructions. By unifying keypoint detection across varied contexts, KptLLM++ establishes itself as an advanced interface, fostering more effective human-AI collaboration. The model is built upon a novel identify-then-detect paradigm, which first interprets keypoint semantics and subsequently localizes their precise positions through a structured chain-of-thought reasoning mechanism.
To push the boundaries of performance, we have scaled up the training dataset to over 500K samples, encompassing diverse objects, keypoint categories, image styles, and scenarios with complex occlusions. This extensive scaling enables KptLLM++ to unlock its potential, achieving remarkable accuracy and generalization. Comprehensive experiments on multiple keypoint detection benchmarks demonstrate its state-of-the-art performance, underscoring its potential as a unified solution for fine-grained image understanding and its transformative implications for human-AI interaction.
\end{abstract}

\section{Introduction}

\begin{figure*}
    \centering
  \includegraphics[width=1\textwidth]{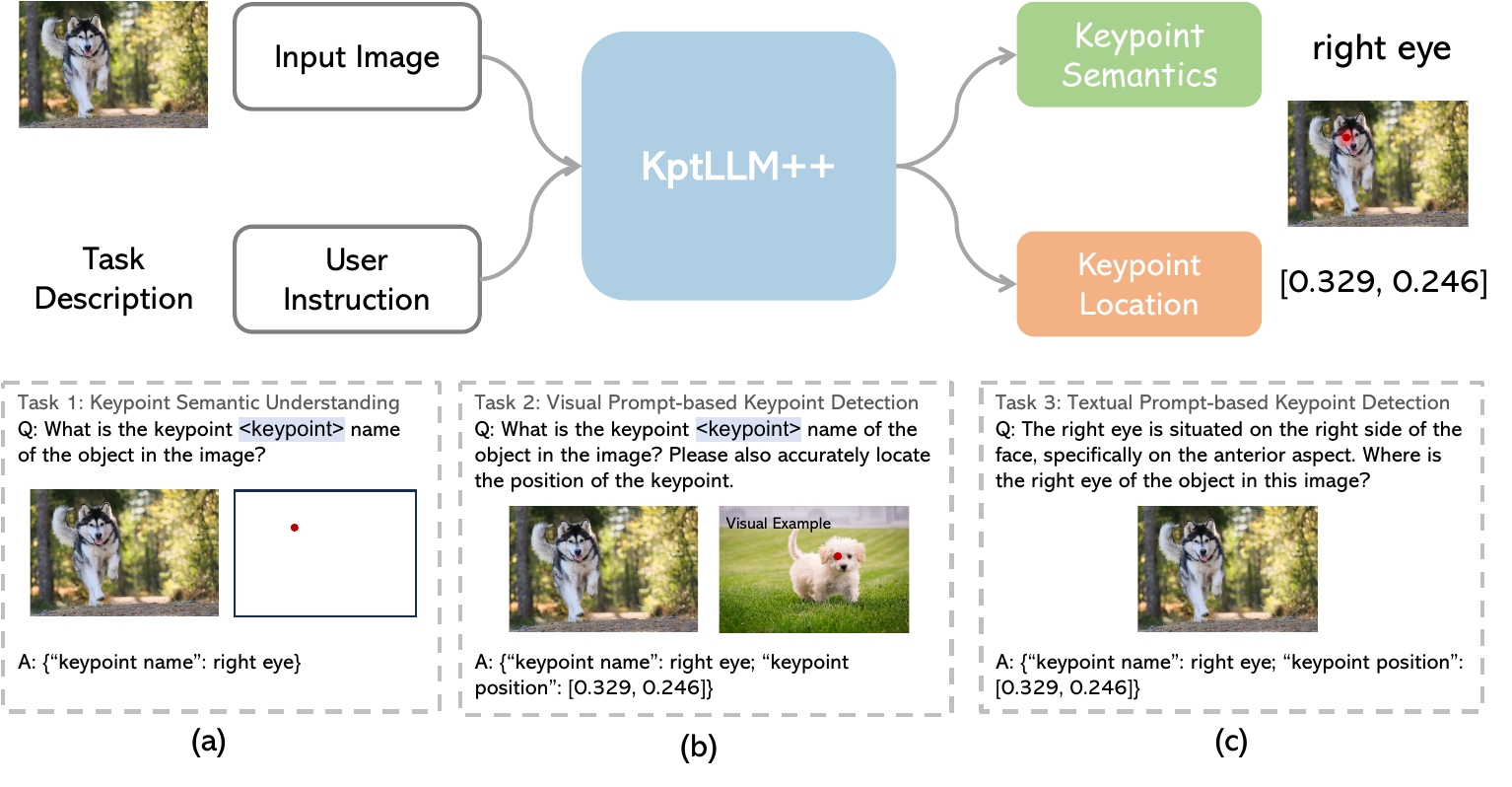}
  \vspace{-0.5cm}
    \caption{This work aims to address the problem of generic keypoint comprehension, which aims to understand keypoints across different task scenarios: \textbf{\textit{(a) Keypoint Semantic Understanding}} takes the object image and a keypoint prompt (\textit{i.e.}, the position of the target keypoint) as inputs, then generate responses that interpret keypoint semantics; \textbf{\textit{(b) Visual Prompt-based Keypoint Detection}} takes a query image and a support image with a keypoint prompt as inputs and then outputs the corresponding keypoint positions and semantics of the query image; \textbf{\textit{(c) Textual Prompt-based Keypoint Detection}} utilizes
detailed descriptions of keypoints through extensive text, to perform more generalizable keypoint detection.
}
\vspace{-0.2cm}
    \label{fig:intro}
\end{figure*}

Recent breakthroughs in deep learning and natural language processing have established Large Language Models (LLMs) as transformative tools for human-computer interaction~\citep{brown2020gpt3, achiam2023gpt4, llama, llama2, chowdhery2023palm, vicuna, mesnard2024gemma}. These models, with their remarkable fluency in text comprehension and generation, serve as bridges connecting users with complex computational tasks. Extending this paradigm, Multimodal Large Language Models (MLLMs)~\citep{li2023blip2, dai2024instructblip, liu2023llava, ye2023mplug, zhu2023minigpt, li2023otter} have been developed to tackle tasks involving both textual and visual modalities, which demonstrated significant success in high-level image understanding. However, they still struggle with fine-grained, pixel-level tasks such as keypoint detection, which require precise semantic and spatial comprehension.

Keypoints represent compact, structure-aware representations of objects and are foundational to a wide range of computer vision applications, including controllable image and video generation~\citep{guo2024liveportrait, zhang2023adding, ju2023humansd}, human-centric perception~\citep{yang2023semantic, yang2023boosting}, and AR/VR systems~\citep{chen2024motionllm, yang2024f, lu2023humantomato}. Traditional methods~\citep{yang2025ed,yang2023explicit} for keypoint detection focus on closed-set problems, aiming to predict predefined semantic keypoints for a specific object category (\textit{e.g.}, human body). However, as the need for generalization grows, researchers explore approaches that leverage visual prompts~\citep{xu2022pose, shi2023matching} or textual prompts~\citep{zhang2023clamp, zhang2024open} to generalize across novel object categories. Despite these advancements, existing methods lack genuine semantic comprehension of keypoints, relying on extensive data fitting rather than understanding keypoint semantics, which leads to misinterpretations and limited adaptability. Additionally, their rigid input-output structures restrict flexibility in user interaction, further limiting their potential as generic solutions for keypoint detection.

To address these limitations, we propose a paradigm that leverages LLMs to enhance keypoint comprehension and unify various keypoint tasks as a whole. By integrating visual and textual modalities, LLMs can serve as intuitive and flexible interfaces, facilitating seamless user interaction and broad generalization. We formalize this paradigm through the problem of \textit{Generic Keypoint Comprehension}, which assesses a model’s ability to understand and localize keypoints across varied scenarios in a semantically meaningful manner. As depicted in Fig.~\ref{fig:intro}, \textit{Generic Keypoint Comprehension} encompasses three interconnected tasks:
\textbf{\textit{(1) Keypoint Semantic Understanding}} infers the semantic meaning of a keypoint based on an image and a descriptive prompt (\textit{i.e.}, keypoint position). This task is critical for applications like structural understanding and behavior recognition.
\textbf{\textit{(2) Visual Prompt-based Keypoint Detection}}, also known as category-agnostic pose estimation, uses a labeled support image to define keypoint semantics, enabling category-agnostic pose estimation and cross-category generalization.
\textbf{\textit{(3) Textual Prompt-based Keypoint Detection}}, also referred to open-vocabulary keypoint detection, leverages detailed natural language descriptions to perform open-vocabulary or zero-shot keypoint detection across arbitrary object and keypoint categories. By addressing these tasks, \textit{Generic Keypoint Comprehension} enhances the versatility of MLLMs and paves the way for generic keypoint detection systems.

Accordingly, our proposed framework, termed KptLLM++, employs an identify-then-detect strategy inspired by human cognition. It first interprets the semantic meaning of keypoints and subsequently localizes them through a structured chain-of-thought reasoning process. By combining commonsense knowledge from LLMs with both visual and textual features, KptLLM++ unifies semantic understanding and precise localization. Specifically, the architecture integrates several carefully designed modules for visual feature extraction, prompt encoding, prompt feature extraction, multi-modal feature fusion via an LLM, and the identify-then-detect strategy to decode both keypoint semantics and locations, effectively handling diverse user-defined tasks. The reasoning process also effectively resolves ambiguities in visually challenging scenarios, such as distinguishing between left and right limbs, aligning predictions with human-like semantic understanding.

To push the boundaries of performance, we have scaled up the training dataset to over 500K samples, covering diverse object types, keypoint categories, image styles, and challenging scenarios with varying occlusions. This extensive scaling empowers KptLLM++ to achieve state-of-the-art performance on various keypoint detection benchmarks and generalize effectively to complex, unseen contexts.
More importantly, KptLLM++ could make keypoint detection a versatile and user-adaptive process, bridging visual and textual domains. It also demonstrates the potential of MLLMs to combine semantic understanding with fine-grained visual analysis, advancing human-AI interaction and setting new standards for keypoint comprehension.

In summary, the contributions of this work are three-fold: \textbf{(1) Generic Keypoint Comprehension:} We pioneer the investigation of this novel problem, aiming to enhance MLLMs with finer-grained keypoint-level image understanding.
\textbf{(2) KptLLM++ Framework:} We introduce a unified multimodal model employing an identify-then-detect strategy to effectively address the three tasks of Generic Keypoint Comprehension, emphasizing initial semantic discernment followed by precise localization through a chain-of-thought process.
\textbf{(3) Superior Performance and Semantic Capabilities:} We scale up the training data to demonstrate KptLLM++'s superiority across various keypoint detection datasets and benchmarks, surpassing state-of-the-art pure vision models as well as vision-language models, while showcasing its unique ability to semantically interpret keypoints.
We hope our work inspires future research in keypoint understanding and localization, while also fostering enhanced human-AI interfaces in fine-grained visual comprehension.
\section{Related Work}

\subsection{Keypoint Detection}
Keypoint detection, also referred to as pose estimation, focuses on localizing the 2D keypoints of objects in an image. This fundamental task has been extensively studied and plays a critical role in various applications, including human-computer interaction, augmented reality, and autonomous systems. Traditional models for keypoint detection are typically designed to handle a single category, such as humans~\citep{lin2014microsoft,mpii,li2019crowdpose,jin2020whole,300w,wu2018look,Moon_2020_ECCV_InterHand2.6M}, animals~\citep{cao2019cross,yu2021ap,khan2020animalweb}, and clothes~\citep{ge2019deepfashion2}. These methods often leverage task-specific priors and architectures tailored to the unique structural and semantic characteristics of the target objects.
Based on the localization strategy, existing keypoint detection methods can be broadly categorized into two groups: regression-based methods and heatmap-based methods. Regression-based methods~\citep{li2021human,nie2019single,sun2017compositional,toshev2014deeppose,yang2023explicit,yang2023neural} predict the precise coordinates of keypoints directly from the input image, often with compact and efficient architectures. In contrast, heatmap-based methods~\citep{chen2018cascaded,cheng2020higherhrnet,jin2019multi,jin2020differentiable,newell2016stacked,sun2019deep,wei2016convolutional,xu2021vipnas,xiao2018simple,yuan2021hrformer,li2021tokenpose,xu2022vitpose} generate heatmaps that represent the likelihood of keypoints at different spatial locations, providing more interpretable and robust predictions.

In recent years, there has been growing interest in methods that generalize beyond predefined categories in the training datasets. One emerging direction is \textbf{visual prompt-based keypoint detection}~\citep{xu2022pose,shi2023matching}, also known as category-agnostic pose estimation. This task involves estimating the pose of any object category in query images using visual prompts, such as a few annotated support images of the novel class. These methods aim to bridge the gap between the traditional fixed-category models and the real-world requirement of handling diverse, unseen object classes.
Another burgeoning area is \textbf{textual prompt-based keypoint detection}~\citep{zhang2023clamp,yang2023unipose}, also referred to open-vocabulary keypoint detection, which aims to localize keypoints based on text prompts in a zero-shot setting. These models exploit the interplay between visual and linguistic representations, enabling them to infer keypoints for arbitrary object categories described in natural language. This approach significantly broadens the applicability of keypoint detection models, allowing them to operate in scenarios with minimal or no annotated training data for the target categories. 
Despite these advancements, current tasks often lack a genuine semantic understanding of keypoints. They typically rely on extensive data fitting to handle keypoint detection rather than comprehending the semantics, which can lead to misinterpretations and limited adaptability. To address this limitation, our work investigates the novel problem of generic keypoint comprehension, which aims to evaluate a model’s ability to understand and localize keypoints across varied scenarios in a semantically meaningful manner.
Accordingly, we propose a unified framework for addressing three distinct but interconnected task scenarios:
(a) keypoint semantic understanding, which focuses on interpreting the semantic meanings of keypoints across various categories.
(b) visual prompt-based keypoint detection, extending from previous tasks, leverages support images and their annotations for novel categories.
(c) textual prompt-based keypoint detection, also rooted in prior tasks, localizes keypoints based on textual descriptions of objects and their keypoints.
This unified framework sets the stage for a comprehensive exploration of keypoint comprehension across diverse settings, significantly advancing the versatility and generalization capabilities of keypoint detection models.

\subsection{Multimodal Large Language Model}
Inspired by the remarkable success of Large Language Models (LLMs)~\citep{brown2020gpt3, achiam2023gpt4, zhang2022opt, llama, llama2, jiang2023mistral, chowdhery2023palm, vicuna, mesnard2024gemma}, researchers are actively exploring ways to adapt and extend the formidable reasoning, comprehension, and generative capabilities of LLMs to the domain of vision. This endeavor has led to the development of Multimodal Large Language Models (MLLMs)~\citep{alayrac2022flamingo, li2023blip2, dai2024instructblip, liu2023llava, liu2023improvedllava, liu2024llavanext, ye2023mplug, li2023otter, lin2023vila, mckinzie2024mm1, lu2024deepseekvl, li2024mgm, laurenccon2024matters, yang2024f}. These models are designed to bridge the gap between natural language understanding and visual perception, unlocking new frontiers in multimodal AI systems.
MLLMs generally adopt an autoregressive paradigm grounded in the transformer decoder architecture~\citep{vaswani2017attention}, which has demonstrated exceptional versatility in processing sequential data. To incorporate visual information, MLLMs typically utilize powerful vision encoders~\citep{radford2021learning, zhai2023sigmoid} to extract features from images or videos. These visual features are subsequently integrated into the LLM framework either through a Multilayer Perceptron (MLP) that maps the features to the LLM's input embedding space~\citep{liu2023llava, liu2023improvedllava, liu2024llavanext}, or via a cross-attention mechanism that dynamically attends to visual content through specialized attention layers~\citep{alayrac2022flamingo, li2023blip2, ye2023mplug}. Such integration enables MLLMs to engage in complex reasoning tasks that require the interplay of visual and linguistic modalities.
Despite these advancements, most existing MLLMs primarily focus on generating textual outputs, which limits their applicability in tasks requiring intricate visual understanding or interaction. For instance, tasks such as object detection, segmentation, and spatial reasoning often require detailed visual perception capabilities that extend beyond text-based descriptions.

To address these challenges, models such as VisionLLM~\citep{wang2024visionllm} have been developed, aiming to handle traditional vision-centric tasks by instruction-tuning LLMs for better alignment with visual tasks. However, while VisionLLM demonstrates enhanced task versatility, it often does not fully harness the comprehensive reasoning capabilities of LLMs for nuanced multimodal applications.
Emerging works, including Kosmos-2~\citep{peng2023kosmos2}, Qwen-VL~\citep{bai2023qwenvl}, F-LMM~\citep{wu2024f}, and DetGPT~\citep{pi2023detgpt}, further extend the utility of LLMs by enabling user-guided detection and grounding through the incorporation of multimodal prompts. These models show promising results in interactive scenarios where users can guide the detection process via natural language or visual instructions. Additionally, models like GPT4RoI~\citep{zhang2023gpt4roi}, Ferret~\citep{you2023ferret}, Shikra~\citep{chen2023shikra} and PerceptionGPT~\citep{pi2023perceptiongpt} introduce innovations by incorporating spatial boxes or masks as inputs, enabling fine-grained region-level visual comprehension. These models are trained on region-text pairs, allowing for a deeper understanding of spatial semantics and improving the granularity of multimodal reasoning.
A notable concurrent work, LocLLM~\citep{wang2024locllm}, specifically explores the use of LLMs for human keypoint localization through textual descriptions. LocLLM demonstrates the feasibility of leveraging LLMs to perform keypoint localization tasks in a text-driven manner, opening new pathways for integrating language and vision in pose estimation.
In contrast, our approach takes a significant step forward by extending the capabilities of MLLMs to comprehend and detect keypoints across diverse object categories using multimodal (\textit{e.g.}, textual or visual) prompts under varying task formulations. By unifying the understanding of keypoints with both visual and textual inputs, our method enhances the interpretative depth and generalization ability of MLLMs. This advancement expands the applicability of MLLMs for keypoint comprehension, enabling them to operate in a wider range of scenarios, from traditional pose estimation to innovative applications that require integrated multimodal reasoning and grounding.

\section{Methodology}
\label{sec:method}

\begin{figure*}
    \centering
\includegraphics[width=0.9\textwidth]{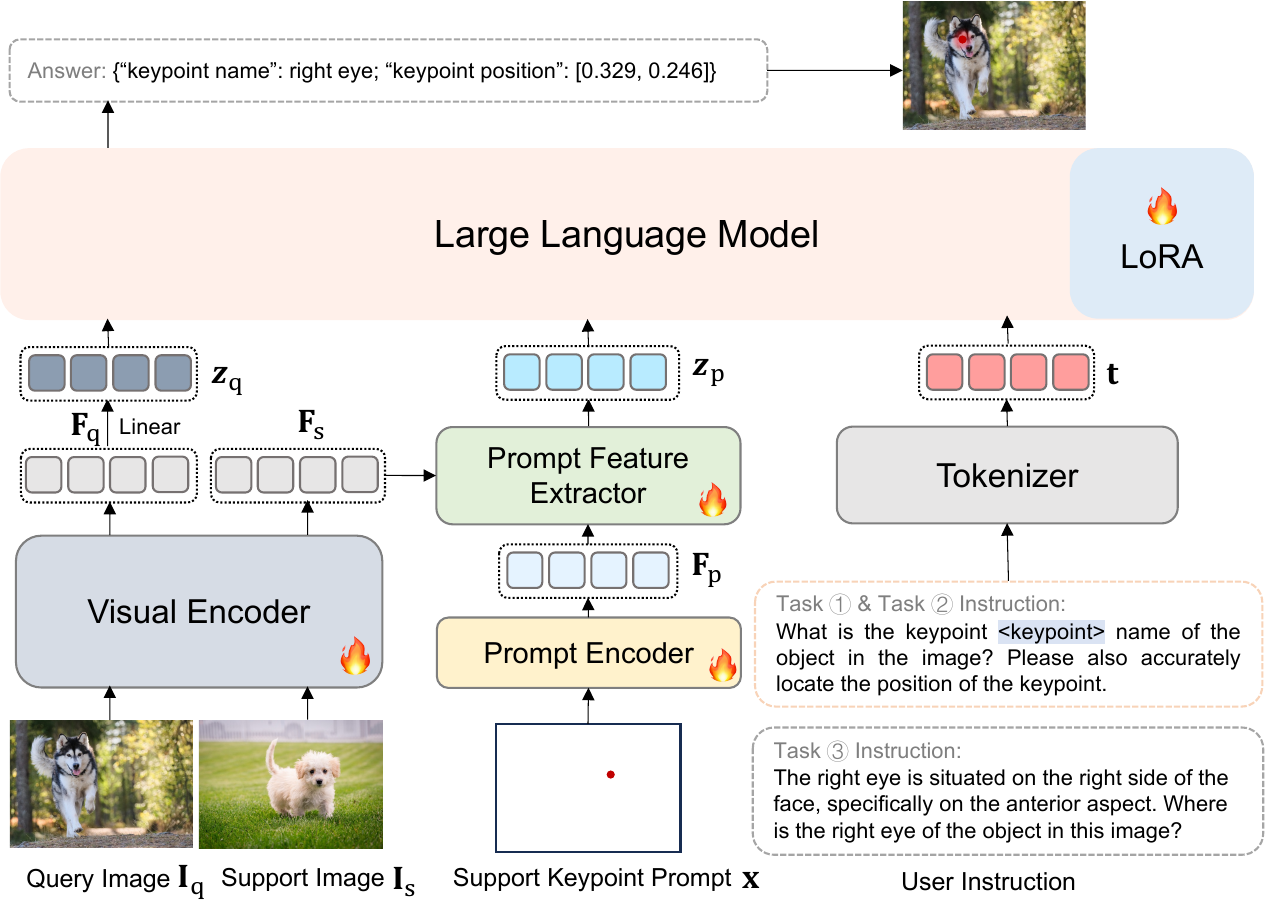}
    \caption{We introduce KptLLM++, a unified framework designed to address three tasks of generic keypoint comprehension:
    \ding{172} \textbf{\textit{Keypoint Semantic Undertanding}}, which processes a support image $\mathbf{I}_s$ and a support keypoint prompt $\mathbf{x}$ to generate responses that interpret the semantic information of the specified keypoint;
    \ding{173} \textbf{\textit{Visual Prompt-based Keypoint Detection}} aims to detect the corresponding keypoint in the query image $\mathbf{I}_q$ based on the understanding of the support keypoint prompt;
    \ding{174} \textbf{\textit{Textual Prompt-based Keypoint Detection}} leverages textual keypoint descriptions to directly infer the corresponding keypoint positions in the query image.
}
% \vspace{-2mm}
    \label{fig:method}
\end{figure*}

This section introduces our proposed unified framework, referred to as KptLLM++, which effectively addresses three generic keypoint comprehension scenarios. 
As illustrated in Fig.~\ref{fig:method}, KptLLM accepts multiple images (query and support images) along with a support keypoint prompt (\textit{i.e.}, the position of the target keypoint in the support image) and textual user instructions as the input. The output comprises both the response text and the desired keypoint position. Specifically,
KptLLM comprises four key architectural components:
(1) A visual encoder that extracts features from both query and support images (see Sec.~\ref{sec:image_encoder});
(2) A prompt encoder that converts support keypoint prompts into prompt embeddings (see Sec.~\ref{sec:Prompt});
(3) A prompt feature extractor that derives prompt-oriented features from the corresponding image features (see Sec.~\ref{sec:Prompt_feature});
(4) A pre-trained LLM that processes multimodal tokens for keypoint comprehension (see Sec.~\ref{sec:LLM}).

\subsection{Visual Encoder}
\label{sec:image_encoder}
The Visual Encoder is designed to process two types of images in parallel: query and support images. Generally, it receives an input image $\mathbf{I} \in \mathbb{R}^{H \times W \times 3}$ and generates a feature map $\mathbf{F} = \mathcal{V}(\mathbf{I}) \in \mathbb{R}^{h \times w \times d}$. Here, $d$ represents the feature dimension, and $h$ and $w$ are the spatial dimensions obtained by downsampling the original image dimensions $H$ and $W$.

\noindent \textbf{Query Image.} The query image represents the image that is to be analyzed. We extract its spatial features through the vision encoder $\mathcal{V}$, resulting in $\mathbf{F}_q$. Following LLaVA~\citep{liu2023llava}, we apply a linear layer to project $\mathbf{F}_q$ into
language space: $\mathbf{z}_q = \texttt{Linear}(\mathbf{F}_q)$. As a result, query visual tokens aligned with the LLM dimension are obtained and fed to the LLM.

\noindent \textbf{Support Image.} The support image serves as a reference example. We extract its spatial features, which are represented as $\mathbf{F}_s$. Unlike the query image features, $\mathbf{F}_s$ is not directly input into LLM. Instead, it is processed by the prompt feature extractor to derive prompt-oriented features.

\noindent \textbf{Choice of Visual Encoder.} The visual encoder can be selected from various options, including ConvNext~\citep{liu2022convnet} and Vision Transformer (ViT)~\citep{vaswani2017attention}. In our preliminary work, KptLLM~\citep{yang2024kptllm}, we employ the ViT-Large from CLIP~\citep{radford2021learning} as the visual encoder. To enhance the perception capabilities of MLLMs, we upgrade to the ViT-Large from DINOv2~\citep{oquab2023dinov2} for KptLLM++, leveraging its superior ability to capture robust visual features.

\subsection{Prompt Encoder}
\label{sec:Prompt}
In addition to processing images, we need to incorporate an additional prompt consisting of 2D coordinates $\mathbf{x} \in \mathbb{R}^{2}$, which describes the keypoint location within the image. 
Inspired by SAM~\citep{kirillov2023segment}, we introduce a prompt encoder to adapt this prompt input to be aligned with the image feature space $\mathbf{F}$.
The prompt encoder encodes the keypoint coordinates using a sine-cosine position embedding (\texttt{PE}), followed by a Multi-Layer Perceptron (\texttt{MLP}):
\begin{equation}
    \mathbf{F}_p =\texttt{MLP}(\texttt{PE}(\mathbf{x})).
\end{equation}

\subsection{Prompt Feature Extractor}
\label{sec:Prompt_feature}

The Prompt Feature Extractor is designed to extract the prompt-specific features from image features. As illustrated in Fig.~\ref{fig:method}, the semantics of the keypoint prompt directly correspond to the support image. We initialize the prompt feature extractor with a two-layer transformer that incorporates the cross-attention mechanism (\texttt{CrossAttnLayers}). This mechanism employs $\mathbf{F}_p$ as the query and $\mathbf{F}_s$ as the key and value to extract keypoint-specific visual features indicated by the prompt:
\begin{equation}
    \mathbf{z}_p =\texttt{CrossAttnLayers}(\mathbf{F}_p,\mathbf{F}_s),
\end{equation}
where $\mathbf{z}_p$ denotes the keypoint-specific visual features. In essence, compared with average pooling-based feature extraction method~\citep{xu2022pose}, the prompt feature extractor is trainable and capable of incorporating global image features to enhance keypoint identification. 
This is particularly beneficial for distinguishing mirror-symmetric keypoints, such as the left and right eyes, which can be highly ambiguous when relying solely on local image features. Our ablation study demonstrates the performance improvements achieved through the utilization of this component.

\subsection{Multimodal LLM for Keypoint Comprehension}
\label{sec:LLM}
Given a query image and an optional prompt specifying the keypoint of interest, our goal is to generate textual descriptions and keypoint locations that convey fine-grained keypoint information within the image.
Recognizing the exceptional ability of LLMs in handling multimodal tokens for different perception tasks~\citep{zang2023contextual, liu2023llava, xu2024pixel, pi2023perceptiongpt, lai2023lisa}, we further leverage LLM for keypoint comprehension, which could effectively process various inputs:
 (1) the visual tokens $\mathbf{z}_q$ of the query image, (2) the prompt tokens $\mathbf{z}_p$, and (3) a sequence of language tokens $\mathbf{t}$, which depend on the three generic keypoint comprehension scenarios.

\noindent \textbf{Keypoint Semantic Decoding.} We design the model to directly generate textual descriptions that interpret keypoint semantics, following the standard approach used by LLMs for text generation. Generally, the architecture of an LLM typically comprises Transformer layers (\texttt{TransformerLayers}) followed by a final Feed Forward Network (\texttt{FFN}). The latent embedding $\mathbf{u}$, which captures the fused multimodal information, can be computed as: \begin{equation} \label{eq
} \mathbf{u} = \texttt{TransformerLayers}([\mathbf{z}_q, \mathbf{z}_p, \mathbf{t}]), \end{equation} where $[\mathbf{z}_q, \mathbf{z}_p, \mathbf{t}]$ denotes the concatenation of the visual, prompt, and language tokens. 
 This embedding $\mathbf{u}$ is then passed through the \texttt{FFN} and a \texttt{Softmax} function to generate the probability distribution $\mathbf{p}$ over the vocabulary for the next token: \begin{equation} \mathbf{p} = \texttt{Softmax}(\texttt{FFN}(\mathbf{u})). \end{equation} 
% where $\mathbf{p}$ denotes the probability distribution over the vocabulary for the next token in the sequence.

\noindent \textbf{Keypoint Position Decoding.} 
Our preliminary work, KptLLM, introduces a special token, \texttt{<keypoint>}, into the vocabulary of the language model. Using this token, the 2D keypoint position $\mathbf{y}$ is computed from the output latent embedding $\mathbf{u}_{\rm{kpt}}$ of the special token via an additional \texttt{FFN} prediction head:
\begin{equation}
        \mathbf{y} = \texttt{FFN}(\mathbf{u}_{\rm{kpt}}) \in \mathbb{R}^{2},
\end{equation}
While effective, this approach requires adding new tokens to the LLM's vocabulary and training an additional FFN prediction head, which reduces simplicity and scalability.

In contrast, KptLLM++ is inspired by~\citep{wang2024locllm}, to directly generate textual descriptions to represent keypoint coordinates, as in Eq.~(4). In this framework, the coordinates are normalized to the range $[0, 1]$ relative to the image size in the spatial dimension, with each value rounded to three decimal places for precision. This textual representation brings multiple advantages: \textit{\textbf{(1) Simplicity:}} By utilizing the existing language generation capabilities of LLMs, KptLLM++ eliminates the need for specialized tokens and additional prediction heads, reducing the architectural complexity.
\textit{\textbf{(2) Scalability:}} The ability to directly generate textual coordinates allows KptLLM++ to scale up efficiently during training, as the model does not require extensive architectural changes to handle diverse keypoint localization tasks across different datasets and scenarios.
\textit{\textbf{(3) Generalizability:}} The textual format for coordinates is versatile and easily interpretable, enabling seamless integration with downstream applications and workflows that benefit from human-readable outputs.

\noindent \textbf{Indentify-then-Detect (ItD) Strategy}.
Instead of relying on extensive data fitting to learn fixed keypoint localization patterns, we adopt an approach where the LLM initially identifies the semantics of keypoints. Subsequently, a chain-of-thought process is employed to accurately detect the locations of these keypoints. Experiments demonstrate improved performance compared to alternative baselines.

\subsection{Training and Inference Details}
\label{sec:training_details}
To retain the learned general knowledge of the pre-trained LLM, we employ LoRA~\citep{hu2021lora} for efficient fine-tuning of LLM, while fully fine-tuning other modules of the framework.
The training and inference processes for different tasks are outlined below.

\noindent \textbf{Keypoint Semantic Understanding.} As shown in Fig.~\ref{fig:intro}-(a),
 this task focuses on extracting semantic textual information associated with specific keypoints within an image. The training objective is to minimize the language modeling loss, computed as the cross-entropy loss over the vocabulary of the LLM's tokenizer. Specifically, the loss function is defined as:
\begin{equation}
\mathcal{L}= \mathcal{L}_{\rm{lm}}(\mathbf{a}, \hat{\mathbf{a}}),
\end{equation}
where $\mathbf{a}$ is the text response predicted by the model, and $\hat{\mathbf{a}}$ is the ground-truth text response.
During inference, given an image provided by the user and the corresponding keypoint position as a prompt, our model comprehends and generates the semantic meaning of the specified keypoint.

\begin{figure*}[t] 
  \centering 
    \includegraphics[width=0.9\textwidth]{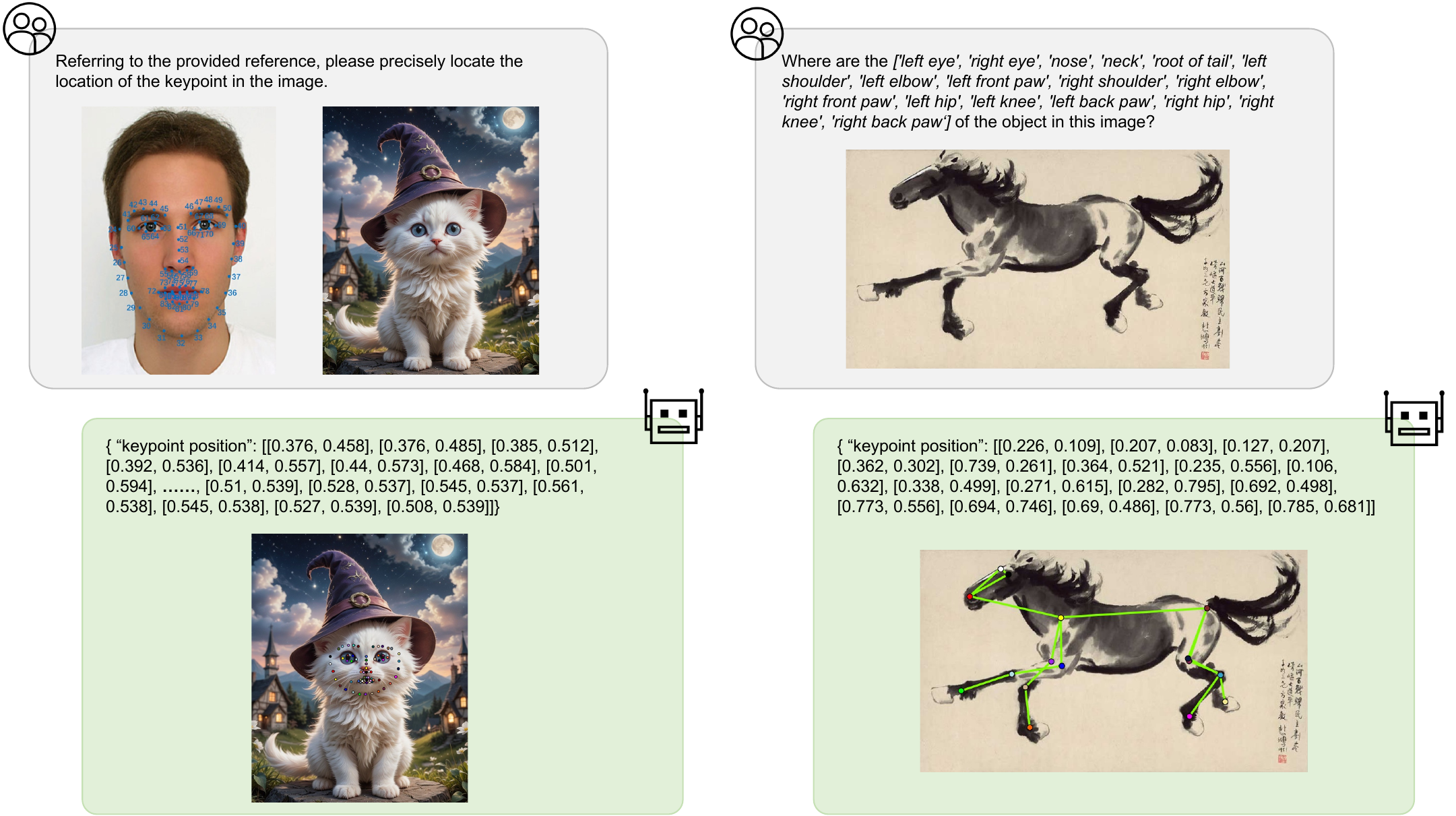}
\caption{User interface of our model for keypoint comprehension under different scenarios. On the left, when a text description is insufficient to describe the target (\textit{e.g.}, dense keypoints on a human face), providing an example as a reference helps the model better understand which keypoints to locate. On the right, explicit keypoint names can be provided to query the model for the corresponding outputs. Note that for better visualization of handling multiple keypoints, we concatenate the queries in the text, but in practice, inference is performed in a batch.}
  \label{fig:interface} 
  \vspace{-0.2cm}
\end{figure*}

\noindent \textbf{Visual Prompt-based Keypoint Detection.} This task involves simultaneously comprehending the semantics of keypoints, generating textual descriptions of this understanding, and precisely localizing the keypoint coordinates, as shown in Fig.~\ref{fig:intro}-(b). 
In our preliminary work KptLLM, the special token-based decoding strategy requires an additional L1 loss for keypoint regression to supervise the prediction of 2D keypoint positions. This loss is combined with the standard language model loss, resulting in the following training objective:
\begin{equation}
\mathcal{L}= \lambda \Vert\mathbf{y}-\mathbf{\hat{y}} \Vert+\mathcal{L}_{\rm{lm}}(\mathbf{a}, \hat{\mathbf{a}})
\end{equation}
where $\mathbf{\hat{y}}$ is the ground-truth keypoint position, and $\lambda$ is the loss weight that balances the learning of keypoint regression and text generation ($\lambda=2$ in our implementation).
In contrast, KptLLM++ adopts a more streamlined approach by leveraging the language model’s ability to directly generate textual descriptions of the keypoint coordinates, thus simplifying the training objective to:
\begin{equation}
\mathcal{L}= \mathcal{L}_{\rm{lm}}(\mathbf{a}, \hat{\mathbf{a}}),
\end{equation}
Here, the unified language model loss encompasses both the generation of textual descriptions and the comprehension of keypoint semantics, ensuring a cohesive training process without the need for specialized components like regression heads.
During inference, the user provides two images: one as the query image for testing and the other as the support image for reference. Additionally, the keypoint definition for the support image should be provided as the support keypoint prompt. Our model then comprehends the semantics of the desired keypoint and detects its corresponding position in the query image.

\noindent \textbf{Textual Prompt-based Keypoint Detection.} As illustrated in Fig.~\ref{fig:intro}-(c), this task aims to accurately localize keypoints based on the detailed keypoint descriptions. 
The loss function aligns with that of the visual prompt-based keypoint detection. 
During inference, users have the option to provide detailed descriptions of the desired keypoints or simply the keypoint names, based on which our model can detect the corresponding keypoints.

\subsection{User Guideline for Keypoint Comprehension}
In practice, KptLLM++ accommodates a variety of user instructions to enable flexible and generic keypoint comprehension. As shown in Fig.~\ref{fig:interface}, the left panel demonstrates that when a text description alone is insufficient to accurately describe the target—such as in cases involving dense keypoints on a human face—providing a visual example as a reference significantly enhances the model’s ability to identify the correct keypoints. This approach is particularly beneficial for complex or ambiguous scenarios where textual descriptions may fall short.
The right panel illustrates an alternative approach, where explicit keypoint names are directly provided as queries to the model. This allows for precise and targeted keypoint extraction, as the model generates outputs corresponding to the specified keypoints. Furthermore, in practice, multiple keypoint queries can be concatenated into a batch, enabling the user to request several keypoints simultaneously for more comprehensive understanding and streamlined processing. This versatility highlights our method’s ability to adapt to different user needs and interaction styles.

\subsection{Real-World Application}
\label{sec:real-world}
\noindent \textbf{Scaling Up.}
The datasets used in our initial work, KptLLM, have limited diversity in object and keypoint categories. While datasets like MP-100~\citep{xu2022pose} and AP-10K~\citep{yu2021ap} are useful, they don’t fully represent real-world scenarios. To improve the model's generalization, we expand the dataset pool by adding eight diverse datasets: COCO~\citep{lin2014microsoft}, Human-Art~\citep{ju2023human}, AP-10K~\citep{yu2021ap}, APT-36K~\citep{yang2022apt}, MacaquePose~\citep{labuguen2021macaquepose}, AnimalWeb~\citep{khan2020animalweb}, Animal Kingdom~\citep{ng2022animal}, and CarFusion~\citep{reddy2018carfusion}. This collection, with over 500,000 samples, enhances diversity in three key areas. \textbf{\textit{Firstly}}, the datasets cover a wide range of keypoint definitions, including human poses, animal anatomy, and objects like vehicles. \textbf{\textit{Secondly}}, they span multiple object categories, from everyday items to more specialized ones. \textbf{\textit{Finally}}, they introduce a variety of instances and contextual diversity, with different poses, backgrounds, and settings like clutter, occlusions, and domain-specific environments. This expanded dataset significantly improves KptLLM++'s ability to generalize, making it more robust and applicable to real-world scenarios.

\begin{figure}[t]
\centering \includegraphics[width=0.4\textwidth]{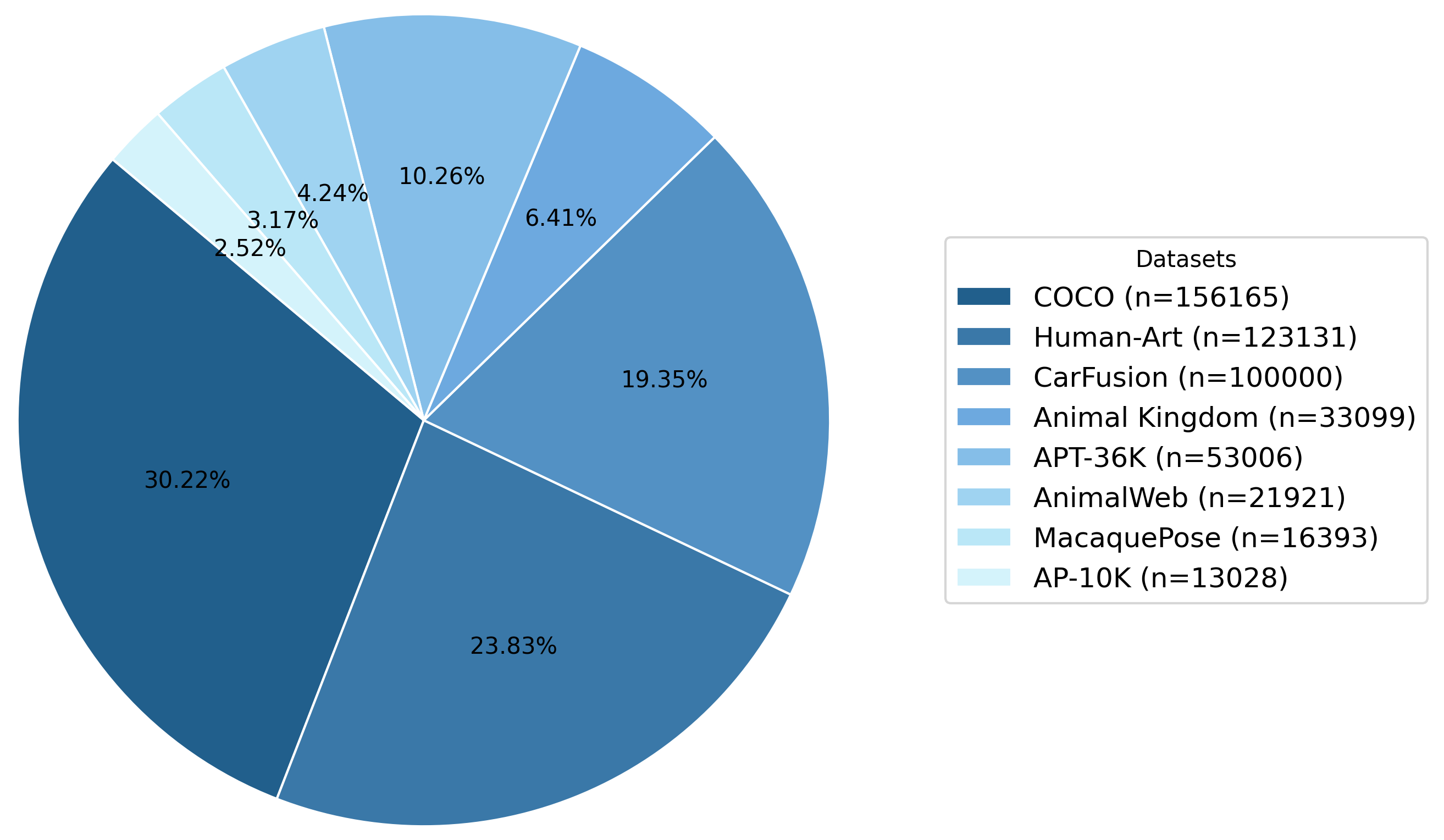}
\caption{Statistical analysis of the datasets used for joint training. Each dataset’s number of instances is denoted as $n$.}
\vspace{-0.2cm}
\label{fig:data}
\end{figure}

\noindent \textbf{End-to-End Pipeline.}
KptLLM++ is initially designed for single-object scenarios, but to extend its applicability to real-world situations involving multiple objects, we combine it with the open-world object detector Grounding-DINO~\citep{liu2025grounding}. The pipeline works as follows: the object detector identifies objects using natural language descriptions and generates bounding boxes. These boxes are then used to crop the image and isolate individual objects. KptLLM++ processes these cropped images to perform precise keypoint localization. By integrating both components, the pipeline enables KptLLM++ to handle complex, multi-object scenarios. The object detector allows for open-world detection based on flexible text descriptions, while KptLLM++ excels at accurately locating keypoints in the isolated objects.

\section{Experiments}

\subsection{Experimental Setup}
\label{sec:experimental_setup}

\subsubsection{Datasets \& Settings} 
In our experiments, we employ different datasets and settings to evaluate the generic keypoint comprehension:

\noindent (1) \textit{\textbf{MP-100 dataset}}\citep{xu2022pose} for \textit{Keypoint Semantic Understanding} and \textit{Visual Prompt-based Keypoint Detection}: This pioneering dataset includes 100 object categories and over 20,000 instances, with keypoints ranging from 8 to 68 per category. It follows the POMNet\citep{xu2022pose} protocol, with five distinct splits: 70 categories for training, 10 for validation, and 20 for testing, ensuring no category overlap for independent training and testing.

\noindent (2) \textit{\textbf{AP-10K dataset}}\citep{yu2021ap} for \textit{Textual Prompt-based Keypoint Detection}: This dataset includes 23 animal families and 54 species with 10,015 images, each annotated with 17 keypoints. We follow CLAMP\citep{zhang2023clamp} to test models’ ability to generalize to unseen species in a zero-shot learning scenario. We design two experimental settings based on whether species are from the same or different animal orders. Species within the same order have similar visual traits, while those from different orders vary more significantly. In the different order setting, Bovidae and Canidae are used for training and testing, while Canidae and Felidae are used for the same order setting.

\begin{table}[t]
    % \vspace{-6mm}
    % \setlength\tabcolsep{8pt}
	\caption{\textbf{Keypoint Semantic Understanding} on MP-100 (Split-1).* means LLaVA is finetuned using LoRA.}
    \vspace{-0.3cm}
	\label{tab:semantic_keypoint}
	\begin{center}
         \resizebox{0.28\textwidth}{!}{
    \begin{tabular}{l|c}
    \hline
    Methods  & Accuracy       \\  \hline  
    LLaVA~\citep{liu2023llava}  & 3\%  \\
    LLaVA$*$~\citep{liu2023llava} & 72\%         \\
    KptLLM & \textbf{83\%} \\
    \hline
    \end{tabular}}
    \vspace{-0.8cm}
    
\end{center}\end{table}

\begin{table*}[t]
\setlength\tabcolsep{6pt}
\caption{\textbf{Visual Prompt-based Keypoint Detection} (PCK) on the MP-100 dataset under 1-shot and 5-shot settings.}
\vspace{-0.3cm}
\label{tab:comparison_example}
\begin{center}
        \resizebox{\textwidth}{!}{
    \begin{tabular}{l|cccccc|cccccc}
        \toprule
        \multirow{2}{*}{Methods} & \multicolumn{6}{c|}{\textbf{1-shot}} &  \multicolumn{6}{c}{\textbf{5-shot}} \\
        & Split1 & Split2 & Split3 & Split4 & Split5 & Mean  & Split1 & Split2 & Split3 & Split4 & Split5 & Mean \\ \hline
        ProtoNet~\citep{snell2017prototypical} & 46.05 & 40.84 & 49.13 & 43.34 & 44.54 & 44.78 & 60.31 & 53.51 & 61.92 & 58.44 & 58.61 & 58.56  \\
        MAML~\citep{finn2017model} & 68.14 & 54.72 & 64.19 & 63.24 & 57.20 & 61.50 & 70.03 & 55.98 & 63.21 & 64.79 & 58.47 & 62.50  \\
    Finetune~\citep{nakamura2019revisiting} & 70.60 & 57.04 & 66.06 & 65.00 & 59.20 & 63.58 & 71.67 & 57.84 & 66.76 & 66.53 & 60.24 & 64.61  \\
        POMNet~\citep{xu2022pose} & 84.23& 78.25 & 78.17 & 78.68 & 79.17 & 79.70& 84.72 & 79.61 & 78.00 & 80.38 & 80.85 & 80.71  \\         CapeFormer~\citep{shi2023matching} & 89.45 & 84.88 & 83.59 & 83.53 & 85.09&  85.31 & 91.94 & 88.92& 89.40 &88.01& 88.25 &89.30\\ \hline
        KptLLM & \textbf{91.66} & \textbf{86.58} & \textbf{86.19} & \textbf{84.76} & \textbf{86.32} & \textbf{87.10}  &  \textbf{93.17} & \textbf{89.45}  & \textbf{90.08} & \textbf{88.74} & \textbf{89.52} & \textbf{90.19}\\
        \bottomrule
    \end{tabular}
    }
    \vspace{-0.3cm}
\end{center}
\end{table*}

\begin{table*}[t]
	\caption{\textbf{Visual Prompt-based Keypoint Detection} for cross super-category evaluation on the MP-100 dataset. Experiments are conducted under the 1-shot setting.}
	\label{tab:cross-category}
    \vspace{-0.3cm}
 % Resize the table to the text width
	\begin{center}
         \resizebox{0.7\textwidth}{!}{
		\begin{tabular}{l|cccc}
			\toprule
			Method & Human Body & Human Face & Vehicle & Furniture \\ \hline
        ProtoNet~\citep{snell2017prototypical} & 37.61 & 57.80 & 28.35 & 42.64 \\
            MAML~\citep{finn2017model} & 51.93 & 25.72 & 17.68 & 20.09 \\
            Fine-tune~\citep{nakamura2019revisiting} & 52.11 & 25.53 & 17.46 & 20.76 \\
            POMNet~\citep{xu2022pose} & 73.82& 79.63 & 34.92 & 47.27 \\
            CapeFormer~\citep{shi2023matching} & 83.44 & 80.96 & 45.40 & 52.49 \\ \hline
            KptLLM & \textbf{83.91} & \textbf{83.37} & \textbf{46.23} & \textbf{54.05}\\
			\bottomrule
		\end{tabular}
		}
	\end{center}
 % \vspace{-0.5cm}
\end{table*}

\noindent (3) For \textbf{\textit{Generic Keypoint Detection}} in \textit{Textual Prompt-based Keypoint Detection}, we jointly train the model using COCO~\citep{lin2014microsoft}, Human-Art~\citep{ju2023human}, AP-10K~\citep{yu2021ap}, APT-36K~\citep{yang2022apt}, MacaquePose~\citep{labuguen2021macaquepose}, AnimalWeb~\citep{khan2020animalweb}, Animal Kingdom~\citep{ng2022animal}, and Carfusion~\citep{reddy2018carfusion}. Specifically, COCO contains over 150,000 human instances with 17 keypoints, while Human-Art focuses on human pose estimation across artistic styles. APT-36K extends AP-10K to a broader range of animal categories, and MacaquePose provides detailed keypoint annotations for macaques. AnimalWeb covers 21.9K images from 334 species, labeled with 9 facial keypoints. Animal Kingdom offers rich animal keypoint annotations, and Carfusion focuses on vehicle keypoint detection. The distribution of the overall training data is shown in Fig.~\ref{fig:data}.

\noindent (4) The \textbf{AnimalPose dataset}~\citep{cao2019cross} serves as the unseen dataset to evaluate generalization for \textit{Textual Prompt-based Keypoint Detection}. It includes animal pose annotations for five categories: dogs, cats, cows, horses, and sheep, with 20 keypoints per instance, covering features like paws, eyes, ears, elbows, nose, and tail base.
\subsubsection{Evaluation and Metrics}
We use different metrics for various datasets and settings to evaluate the generic keypoint comprehension capability.
(1) On the \textbf{\textit{MP-100 dataset}}, for \textit{Keypoint Semantic Understanding}, we adopt the keypoint semantic labels from X-Pose~\citep{yang2023unipose}. Some keypoints, like those describing the collar in the clothing category, are excluded due to ambiguity. We aggregate the results to derive accuracy rates (%). For \textit{Visual Prompt-based Keypoint Detection}, we use the Probability of Correct Keypoint (PCK) metric, with a uniform threshold of 0.2 across all categories, as in POMNet~\citep{xu2022pose}. The average PCK is computed across all five data splits for a comprehensive evaluation.
(2) On the \textit{\textbf{AP-10K dataset}}\citep{yu2021ap}, we follow CLAMP and use average precision (AP), computed based on object keypoint similarity (OKS), as the primary metric.
(3) For \textbf{\textit{Generic Keypoint Detection}}, we apply standard metrics such as AP, Average Recall (AR), and PCK, with dataset-specific thresholds.
(4) On the \textbf{\textit{unseen AnimalPose dataset}}\citep{cao2019cross}, we use PCK@$0.2$ following X-Pose~\citep{yang2023unipose} to assess the model's generalization performance.

\subsubsection{Implementation Details}
\label{sec:implementation_details}

\textbf{Architecture.} We implement two model versions: KptLLM and KptLLM++. The first model, KptLLM, utilizes LLaVA-V1.5-7B~\citep{liu2023llava} as the base model, which incorporates the ViT-based visual encoder from CLIP for image encoding and Vicuna-7B (fine-tuned from Llama-2) as the LLM backbone. We apply LoRA for efficient fine-tuning of the LLM, while all other modules, including the visual encoder, prompt encoder, prompt feature extractor, and a series of linear layers and feedforward networks, undergo full fine-tuning. The input image contains a single object of interest, cropped based on the ground-truth bounding box and resized to 336$\times$336, in line with CLIP-ViT-L specifications.
The second model, KptLLM++, uses the ViT-based visual encoder from DINOv2 and Vicuna-7B as the LLM. LoRA is used for efficient fine-tuning of both the LLM and visual encoder, while other modules are fully fine-tuned. The input image, which also contains a single object, is resized to 224$\times$224, following~\citep{wang2024locllm}.

\noindent \textbf{Training Details.} LoRA parameters are configured with a rank of 128 and an alpha of 256. Optimization is conducted using AdamW, with a learning rate of 2e$-$4 and weight decay of 0. We utilize 8 NVIDIA A100 GPUs for training. Each GPU operates with a batch size of 16, and we employ a gradient accumulation step of 1.

\begin{figure*}[h]
    \centering
\includegraphics[width=0.9\textwidth]{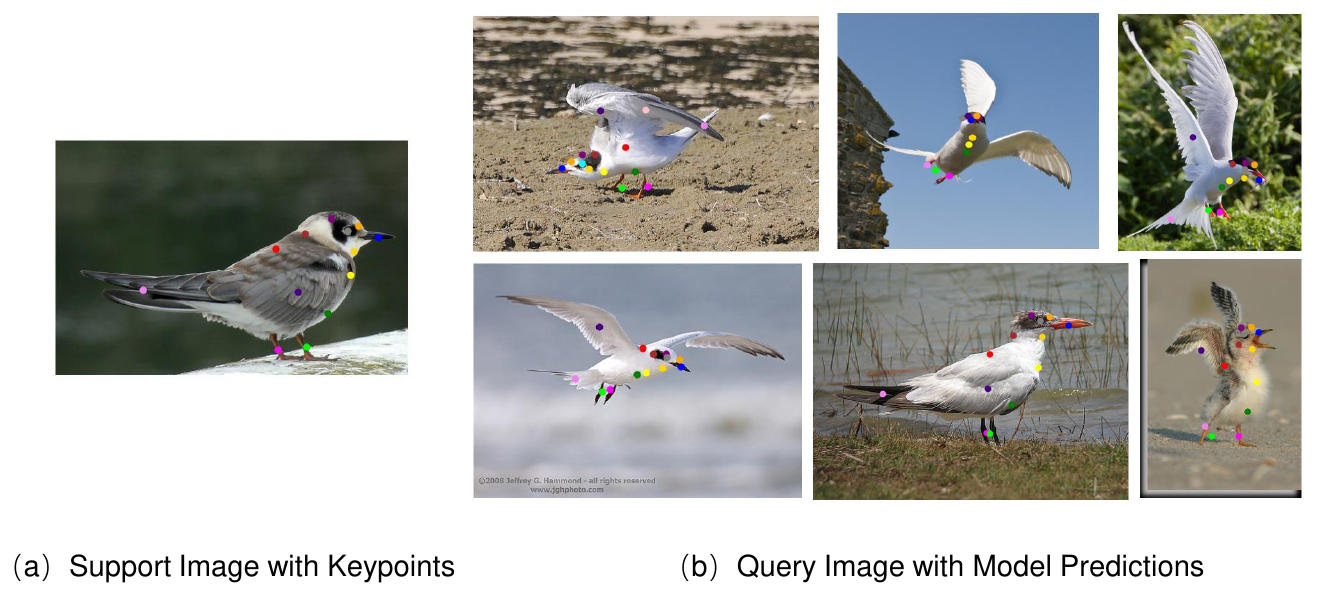}
\vspace{-0.1cm}
    \caption{Using the same support image with support keypoints, our model could effectively detect different query images with various poses, object appearances, and environments.
}
    \label{fig:mp100_vis}
% \vspace{-0.3cm}
\end{figure*}

\begin{table*}[t]
  \centering
    \caption{\textbf{Textual Prompt-based Keypoint Detection} on the AP-10K dataset~\citep{yu2021ap}.}
    \vspace{-0.3cm}
             \resizebox{0.9\textwidth}{!}{
  \begin{tabular}{l|c|c|ccc ccc}
    \toprule
    Method  & Train & Test & $AP$ & $AP_{50}$ & $AP_{75}$ & $AP_{M}$ & $AP_{L}$ & $AR$ \\
    \hline
    SimpleBaseline~\citep{xiao2018simple} & Bovidae & Canidae & 41.3 &	79.4 & 36.4	& 26.8 & 41.3 &	49.1\\
    CLAMP~\citep{zhang2023clamp} & Bovidae & Canidae & 46.9 &	84.4 & 45.6 & \textbf{30.3} & 46.9 &	53.8\\
    KptLLM & Bovidae & Canidae& \textbf{62.2} &  \textbf{94.5} & \textbf{68.7} & 26.0 &\textbf{62.3} & \textbf{65.8} \\
    \hline
    SimpleBaseline~\citep{xiao2018simple}  & Canidae & Felidae & 39.6 & 74.1 & 34.5 & 9.5 & 40.2 & 46.6\\
    CLAMP~\citep{zhang2023clamp}  & Canidae & Felidae & 48.4 & 85.7 & 44.0 & 13.6 & 48.9 & 55.1\\
        KptLLM & Canidae & Felidae & \textbf{70.2} &  \textbf{97.8} & \textbf{82.2} & \textbf{49.2} &\textbf{70.6} & \textbf{74.7}\\
    \bottomrule
  \end{tabular}}
  \label{ap10k-zero-shot}
  % \vspace{-0.5cm}
\end{table*}

\begin{table*}[t]
  \centering
    \caption{Comparison of KptLLM and KptLLM++.}
    \vspace{-0.3cm}
             \resizebox{0.9\textwidth}{!}{
  \begin{tabular}{l|c|c|c|c|c}
    \toprule
    Method  & Visual Encoder  & Decoding Strategy  & Scaling Up & COCO (seen) & AnimalPose (unseen) \\
    \hline
    KptLLM~\citep{yang2024kptllm} & CLIP  & Feature Regression  & $\times$& 76.2 & 55.9 \\
    KptLLM++ & DINOv2  & Numerical Text & $\times$& 77.6 & 59.1\\
    KptLLM++ & DINOv2 &  Numerical Text  & $\checkmark$  & \textbf{78.1} & \textbf{79.2} \\
    \bottomrule
  \end{tabular}}
  \label{tab:kptllm_vs_kptllm++}
  % \vspace{-0.5cm}
\end{table*}

\begin{figure*}[t] 
  \centering 
\includegraphics[width=1\textwidth]{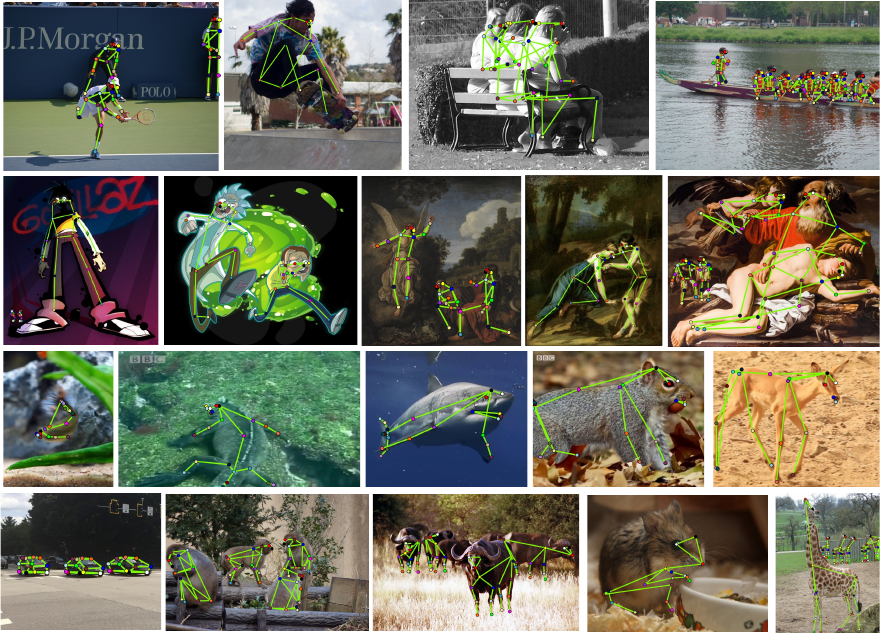}
  \caption{Visualization Results of KptLLM++, including COCO, Human-Art, AP-10K, MacaquePose, Animal Kingdom and CarFusion. We use dataset-specific keypoint descriptions to query the model for keypoint detection.} 
  \label{fig:vis_indomain} 
  % \vspace{-0.2cm}
\end{figure*}

\begin{figure*}[t] 
  \centering 
    \includegraphics[width=1\textwidth]{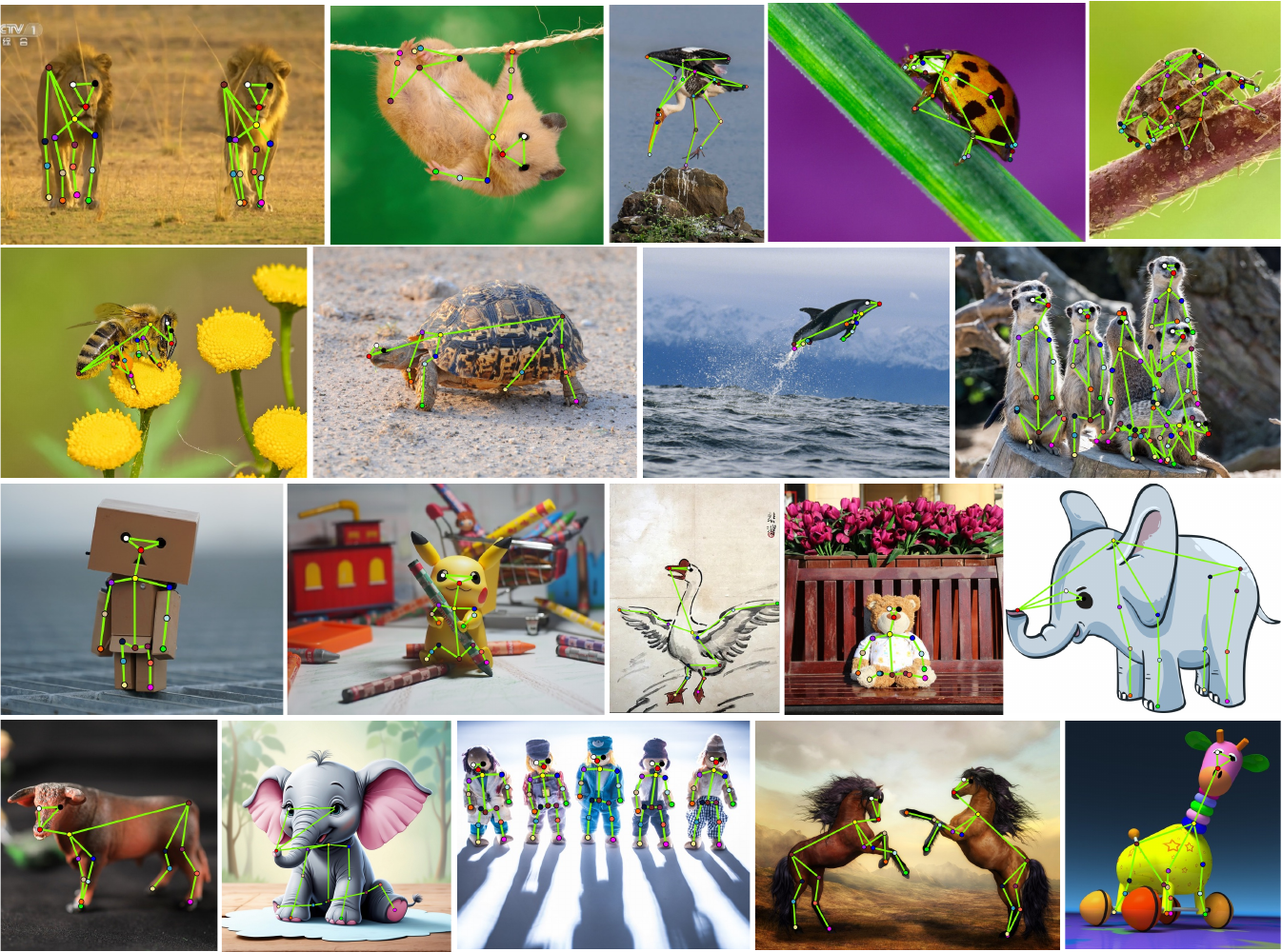}
  \caption{
  % Visualization Results of KptLLM++ for Arbitrary Images on the Internet via textual prompts, e.g., the predefined keypoint definitions in the AP-10K dataset. 
  Visualization results of KptLLM++ via \textit{Textual Prompt-based Keypoint Detection} on the images from the Internet containing arbitrary categories.
  As illustrated in Sec.~\ref{sec:real-world}, we employ the open-set object detector-GroundingDINO~\citep{liu2025grounding} to provide boundingbox of the desired object and further use KptLLM++ to detect the corresponding keypoints.} 
  \label{fig:ood_test} 
  % \vspace{-0.2cm}
\end{figure*}

\subsection{Keypoint Semantic Understanding} 

As depicted in Tab.~\ref{tab:semantic_keypoint}, we present the accuracy for keypoint semantic understanding on MP-100~\citep{xu2022pose} Split-1 set. 
To facilitate a comprehensive comparison, we highlight the keypoint area in the image and feed the processed image, along with the task instruction, into LLaVA~\citep{liu2023llava}. We report the performance of both the original LLaVA model and a version fine-tuned on the MP-100 dataset. 
The original LLaVA performs notably poorly in grasping keypoint semantics, indicating the inadequacy of traditional multimodal large language models in capturing fine-grained semantic details. 
Conversely, the fine-tuned LLaVA demonstrates significantly enhanced performance, thereby validating the efficacy of our training pipeline. Furthermore, our KptLLM surpasses the fine-tuned LLaVA by a substantial margin, particularly in terms of keypoint accuracy (83\% vs 72\%). It demonstrates the effectiveness of our keypoint prompt token in guiding attention to the fine-grained keypoint area.

\subsection{Visual Prompt-based Keypoint Detection}
\textbf{1-shot and 5-shot Evaluation}. We compare our method with previous visual prompt-based methods like ProtoNet~\citep{snell2017prototypical}, MAML~\citep{finn2017model}, Fine-tune~\citep{nakamura2019revisiting}, POMNet~\citep{xu2022pose}, and CapeFormer~\citep{shi2023matching}. Tab.~\ref{tab:comparison_example} shows the PCK results on the MP-100 dataset for both 1-shot and 5-shot settings. KptLLM consistently outperforms these methods across all settings and data splits, demonstrating the effectiveness of MLLM for keypoint detection. Additionally, we incorporate keypoint semantic understanding, offering new capabilities for interpreting the semantic aspects of support image keypoints.

\noindent \textbf{Cross Super Category Evaluation.} 
To thoroughly assess generalization across markedly different categories, we conduct a cross-supercategory evaluation following the protocol of POMNet~\citep{xu2022pose}. While the MP-100 dataset ensures that training, validation, and test categories are non-overlapping, some categories may still exhibit similar features, \textit{e.g.}, body characteristics commonly shared among different quadruped animals. To address this, we designate four supercategories—human face, human body, vehicle, and furniture—from the MP-100 dataset as test categories. The remaining categories are utilized for training, allowing us to better evaluate the model's ability to generalize across significantly diverse categories.
As shown in Tab.~\ref{tab:cross-category}, KptLLM consistently outperforms previous methods, highlighting the robustness and excellent generalization ability of our proposed method.

\noindent \textbf{Qualitative Results.} 
For the visual prompt-based keypoint detection task, the input necessitates a support image of the object to be tested, as well as keypoint positions that represent the definitions of those keypoints. Here, we examine how the visual disparity between support images and query images affects the model's performance. As in Fig.~\ref{fig:mp100_vis}, our model is capable of effectively detecting keypoints in various query images when provided with the same support image and its corresponding keypoints. This effectiveness is maintained even in the presence of differences in object poses, appearances, and environmental conditions.

\subsection{Textual Prompt-based Keypoint Detection}

The results are presented in Tab.~\ref{ap10k-zero-shot}. Compared to previous textual prompt-based model-CLAMP~\citep{zhang2023clamp}, KptLLM achieves superior cross-species generalization. Specifically, our model shows a $15.3$ AP improvement in the different order setting and a $21.8$ AP increase in the same order setting. Notably, our model performs better in the same order setting, where species often share similar visual characteristics. Overall, we show that leveraging detailed keypoint descriptions, combined with commonsense knowledge from LLMs, effectively enhances generalizable performance in keypoint localization.

\subsection{Generic Keypoint Detection}
\textbf{Comparison of KptLLM and KptLLM++.} As detailed in Sec.~\ref{sec:method}, KptLLM++ has refined the model design in our preliminary work KptLLM~\citep{yang2024kptllm}. Specifically, we have improved the decoding strategy used in KptLLM, which employs hidden features of the special token and feed-forward networks (FFNs) for regressing keypoint positions, by directly outputting keypoint coordinates as textual descriptions. This adjustment enhances the model's scalability. Additionally, we replace the ViT-based visual encoder of CLIP used in KptLLM with the visual encoder from DINO-v2. As shown in Tab.~\ref{tab:kptllm_vs_kptllm++}, we compare KptLLM and KptLLM++ across various aspects. The ablation study reveals that KptLLM++ not only exhibits significantly better performance on in-domain tests (\textit{e.g.}, the seen dataset COCO~\citep{lin2014microsoft}) but also demonstrates exceptional generalization capabilities on out-of-domain tests (\textit{e.g.}, the unseen dataset AnimalPose~\citep{cao2019cross}). 

\noindent \textbf{Comparison of KptLLM++ and the SOTA Methods on Various Keypoint Datasets.} 
Training a model with a single dataset that has limited diversity in object and keypoint categories significantly reduces its applicability, making it insufficient for handling open-world scenarios. To probe the upper bounds of MLLMs in keypoint comprehension, we conduct a scaling-up experiment through joint training with multiple datasets. Specifically, we employ eight large-scale datasets for training, including COCO~\citep{lin2014microsoft}, Human-Art~\citep{ju2023human}, AP-10K~\citep{yu2021ap}, APT-36K~\citep{yang2022apt}, MacaquePose~\citep{labuguen2021macaquepose}, AnimalWeb~\citep{khan2020animalweb}, Animal Kingdom~\citep{ng2022animal} and Carfusion~\citep{reddy2018carfusion}.
For evaluation, we compare its performance with other state-of-the-art methods, including vision-only models and vision-language models. The details are as follows: \textbf{\textit{(1) COCO.}} We compare KptLLM++ with other methods on COCO \texttt{val} set. As shown in Tab.~\ref{tab:coco}, the comparison includes vision-only models, such as several representative and state-of-the-art methods like ViTPose~\citep{xu2022vitpose} and HRNet~\citep{sun2019deep}, as well as vision-language models, including the recent keypoint generalist X-Pose~\citep{yang2023unipose}, the LLM-based models VisionLLM v2~\citep{wu2024visionllm}, and LocLLM~\citep{wang2024locllm}. Compared to all these methods, KptLLM++ achieves the best performance, breaking the previously perceived performance upper bound for LLM-based approaches.
\textbf{\textit{(2) Human-Art.}} Beyond natural human-centric scenes, KptLLM++ also demonstrates outstanding performance in artistic scenarios, as in Tab.~\ref{tab:humanart}. For example, it achieves 1.4 AP higher than X-Pose and 4.7 AP higher than VisionLLM v2.
\textbf{\textit{(3) MacaquePose.}} For humanoid animals, such as macaques, KPTLLM++ also achieves state-of-the-art performance, with an AP of 83.9, as shown in Tab.~\ref{tab:MacaquePose}.
\textbf{\textit{(4) AP-10K.}} For 23 animal families and 54 species, KptLLM++ achieves state-of-the-art performance, surpassing X-Pose by 2 AP and VisionLLM v2 by 3.4 AP, as illustrated in Tab.~\ref{tab:ap10k}.
\textbf{\textit{(5) Animal Kingdom.}}
For more diverse animals, KptLLM++ consistently achieves state-of-the-art performance, as shown in Tab.~\ref{tab:ak}.
\textbf{\textit{(6) CarFusion.}}
For rigid structures, such as cars, KptLLM++ also outperforms all previous methods, as shown in Tab.~\ref{tab:carfusion}. 

\noindent \textbf{Comparison of KptLLM++ and the SOTA Method on the Unseen Dataset.}
We conduct generalization validation experiments following X-Pose~\citep{yang2023unipose} to evaluate KptLLM++ on the unseen dataset AnimalPose~\citep{cao2019cross}. As shown in Tab.~\ref{tab:generalizable}, KptLLM++ demonstrates superior generalization compared to X-Pose. Furthermore, as observed in Tab.~\ref{tab:kptllm_vs_kptllm++}, we find that scaling up the training data significantly enhances generalization performance. This highlights the importance of training the model with a more diverse set of keypoint definitions.

\noindent \textbf{Qualitative Results.} As shown in Fig.~\ref{fig:vis_indomain}, KptLLM++ demonstrates strong keypoint localization capabilities across in-domain datasets, including COCO, Human-Art, AP-10K, MacaquePose, Animal Kingdom, and CarFusion.

\begin{table*}[t]
	\caption{\textbf{Generic Keypoint Detection} on the COCO \texttt{val}. All top-down methods utilize ground-truth (GT) bounding boxes for testing. All results are obtained without employing flip tests.}
        \vspace{-0.3cm}
	\label{tab:coco}
 % Resize the table to the text width
	\begin{center}
         \resizebox{0.7\textwidth}{!}{
		\begin{tabular}{l|ccc ccc}
			\toprule
			Method  & $AP$ & $AP_{50}$ & $AP_{75}$ & $AP_{M}$ & $AP_{L}$ & $AR$ \\ \hline
            \rowcolor{Gray!80} \multicolumn{7}{l}{\emph{Vision-only Models}} \\
            SimplePose~\citep{xiao2018simple} &   74.4 & 92.6 & 82.5 & 71.5&  79.2 & 77.6\\
            HRNet~\citep{sun2019deep} &  76.8 & 93.6 & 83.6 & 74.0 & 81.5 & 79.6\\
            SimCC~\citep{li2022simcc} &  76.5 & 93.2 & 83.1 & 73.6 & 81.5 & 79.7 \\
            ViTPose~\citep{xu2022vitpose} & 77.4 & 93.6 & 84.8 & 74.7 & 81.9 & 80.2 \\ \hline
            \rowcolor{Gray!80} \multicolumn{7}{l}{\emph{Vision-Language Models}} \\
            X-Pose~\citep{yang2023unipose} &  76.8 & 91.9 & 83.8 & 71.6 & 84.8 & -\\
            VisionLLM v2~\citep{wu2024visionllm}  & 74.2 & - & - & - & - & - \\
            LocLLM~\citep{wang2024locllm} & 77.4 & \textbf{94.4} &  85.2 & 74.5 & 81.8 & 80.6\\
            KptLLM++ & \textbf{78.1} & 94.3 & \textbf{85.3} & \textbf{75.0} & \textbf{83.1} & \textbf{81.3} \\
			\bottomrule
		\end{tabular}
		}
	\end{center}
\end{table*}

\begin{table*}[t]
	\caption{\textbf{Generic Keypoint Detection} on the AP-10K \texttt{val}. All top-down methods utilize GT bounding boxes for testing.}
        \vspace{-0.3cm}
	\label{tab:ap10k}
 % Resize the table to the text width
	\begin{center}
         \resizebox{0.65\textwidth}{!}{
		\begin{tabular}{l|ccc cc}
			\toprule
			Method  & $AP$ & $AP_{50}$ & $AP_{75}$ & $AP_{M}$ & $AP_{L}$ \\ \hline
            \rowcolor{Gray!80} \multicolumn{6}{l}{\emph{Vision-only Models}} \\
                SimplePose~\citep{xiao2018simple} &  68.1 & 92.2 & 74.2 & 53.4 & 68.8\\
            HRNet~\citep{sun2019deep} &  73.1 & 93.7 &  80.4 & 57.4 & 73.8\\
            ViTPose~\citep{xu2022vitpose} & 80.1 & 97.5 & 88.0 & 62.3 & 80.3 \\ \hline
            \rowcolor{Gray!80} \multicolumn{6}{l}{\emph{Vision-Language Models}} \\
            X-Pose~\citep{yang2023unipose} & 79.2 & 95.7 & 87.2 & 58.3 & 79.8 \\
            VisionLLM v2~\citep{wu2024visionllm}  & 77.8 & - & - & -  & - \\
            KptLLM++ & \textbf{81.2} & \textbf{98.1} & \textbf{88.6} & \textbf{60.9} & \textbf{81.9} \\
			\bottomrule
		\end{tabular}
		}
	\end{center}
         \vspace{-0.3cm}
\end{table*}

\begin{table*}[t]
	\caption{\textbf{Generic Keypoint Detection} on the Human-Art \texttt{val}.}
        \vspace{-0.3cm}
	\label{tab:humanart}
 % Resize the table to the text width
	\begin{center}
         \resizebox{0.7\textwidth}{!}{
		\begin{tabular}{l|ccc ccc}
			\toprule
			Method  & $AP$ & $AP_{50}$ & $AP_{75}$ & $AP_{M}$ & $AP_{L}$ & $AR$ \\ \hline
            \rowcolor{Gray!80} \multicolumn{7}{l}{\emph{Vision-only Models}} \\
            SimplePose~\citep{xiao2018simple} &   48.4 &73.0 & 50.7 & 27.2 & 50.7 & 52.8\\
            HRNet~\citep{sun2019deep} &  51.7 & 75.2 & 54.8 & 26.2 & 54.3 & 57.0\\
            SimCC~\citep{li2022simcc} &  53.4 & 76.3 & 56.5 & 30.4 & 55.9 & 57.5\\
            ViTPose~\citep{xu2022vitpose} & 53.8 & 77.9 &  57.4 & 31.4&  56.6&  58.7\\ \hline
            \rowcolor{Gray!80} \multicolumn{7}{l}{\emph{Vision-Language Models}} \\
            X-Pose~\citep{yang2023unipose} & 75.9 & 89.6 & 81.7 & 42.6 & 80.1 & -\\
            LocLLM~\citep{wang2024locllm} & 64.8 & 87.4 & 70.4 & 40.9 & 67.4 & 69.3 \\
            VisionLLM v2~\citep{wu2024visionllm}  & 72.6 & - & - & - & - & - \\
            KptLLM++ & \textbf{77.3} & \textbf{92.2} & \textbf{83.1} & \textbf{54.7} & \textbf{80.1} &\textbf{80.6}\\
			\bottomrule
		\end{tabular}
		}
	\end{center}
         \vspace{-0.3cm}
\end{table*}

\begin{table}[t]
	\caption{\textbf{Generalizable Keypoint Detection} on the unseen dataset-AnimalPose.}
        \vspace{-0.3cm}
	\label{tab:generalizable}
	\begin{center}
         \resizebox{0.25\textwidth}{!}{
		\begin{tabular}{l|c}
			\toprule
			Method   & PCK$@$0.2 \\ \hline
            X-Pose~\citep{yang2023unipose} & 73.4 \\
            KptLLM++ & \textbf{79.2} \\
			\bottomrule
		\end{tabular}
		}
	\end{center}
    \vspace{-0.5cm}
\end{table}

\begin{table*}[ht]
    \begin{center}
    \begin{minipage}{0.3\linewidth}
        \caption{{\textbf{Generic Keypoint Detection} on MacaquePose.}}
        \setlength{\tabcolsep}{15pt}
        \vspace{-0.3cm}
        \centering
        \label{tab:MacaquePose}
            % \setlength\tabcolsep{20pt}
            % \makeatletter\def\@captype{table}\makeatother
                \begin{tabular}{l|c}
             \toprule
                Method   & AP \\
                \hline
                            \rowcolor{Gray!80} \multicolumn{2}{l}{\emph{Vision-only Models}} \\
                ViTPose &  15.5 \\
                % ED-Pose~\citep{yang2023explicit} & \\
                        \rowcolor{Gray!80} \multicolumn{2}{l}{\emph{Vision-Language Models}} \\
                X-Pose &  79.4\\    
                    VisionLLM v2& 83.1 \\    
                KptLLM++ & \textbf{83.9} \\
       \bottomrule
                \end{tabular} 
    \end{minipage}
    \quad
    \begin{minipage}{0.3\linewidth}
        \caption{{\textbf{Generic Keypoint Detection} on Animal Kingdom.}}
                \vspace{-0.3cm}
        \label{tab:ak}
        \centering
        \setlength{\tabcolsep}{6pt}
                    % \setlength\tabcolsep{20pt}
            % \makeatletter\def\@captype{table}\makeatother
            \begin{tabular}{l|cc}
    \toprule
            Method      & PCK$@$0.05  & PCK$@$0.2        \\
            \hline
            \rowcolor{Gray!80} \multicolumn{3}{l}{\emph{Vision-only Models}} \\
            HRNet& 63.23   & -      \\
        \rowcolor{Gray!80} \multicolumn{3}{l}{\emph{Vision-Language Models}} \\
        X-Pose& - &  \textbf{96.1} \\
        VisionLLM v2 & - & 95.5 \\
            KptLLM++ & \textbf{69.64} & \textbf{96.1}\\
            \bottomrule
            \end{tabular}
    \end{minipage}
    \quad
\begin{minipage}{0.28\linewidth}
    \caption{{\textbf{Generic Keypoint Detection} on CarFusion}.}
            \vspace{-0.1cm}

    \label{tab:carfusion}
        \centering
                    % \setlength\tabcolsep{20pt}
        % \makeatletter\def\@captype{table}\makeatother
        \begin{tabular}{l|cc}
        \toprule
        Method         &  AP    & PCK$@$0.2           \\
    \rowcolor{Gray!80} \multicolumn{3}{l}{\emph{Vision-only Models}} \\
        \hline
        ED-Pose   &  64.7 & -          \\
        \rowcolor{Gray!80} \multicolumn{3}{l}{\emph{Vision-Language Models}} \\
        X-Pose & - & 88.7 \\
        % VisionLLM v2~\citep{wang2024visionllm} & - &  \\
        KptLLM++    &  \textbf{79.8} &  \textbf{99.4} \\
        \bottomrule
        \end{tabular}
\end{minipage}
\end{center}
% \vspace{-0.2cm}
\end{table*}

\begin{table}[ht]
    \begin{center}
    \begin{minipage}{0.45\linewidth}
        \caption{{Ablation study of semantic understanding.}}
        \vspace{-0.3cm}
        \centering
        \label{tab:ablation_semantic}
            \setlength\tabcolsep{10pt}
            \makeatletter\def\@captype{table}\makeatother
                \begin{tabular}{c|c}
             \toprule
                Strategies   & PCK \\
                \hline
                w/o ItD & 87.68 \\
                w/ ItD  & \textbf{91.66} \\
       \bottomrule
                \end{tabular} 
    \end{minipage}
    \quad
    \begin{minipage}{0.45\linewidth}
        \caption{{Ablation study of prompt feature extraction.}}
                \vspace{-0.3cm}
        \label{tab:ablation_prompt}
        \centering
            \setlength\tabcolsep{10pt}
            \makeatletter\def\@captype{table}\makeatother
            \begin{tabular}{l|c}
    \toprule
            Strategies      & PCK           \\
            \hline
            Average Pooling & 89.78         \\
            Ours & \textbf{91.66}           \\
            \bottomrule
            \end{tabular}
    \end{minipage}
    \quad
\begin{minipage}{0.45\linewidth}
    \caption{{Ablation study of visual and textual prompts.}}
            \vspace{-0.3cm}
    \label{tab:ablation_two_prompt}
        \centering
\makeatletter\def\@captype{table}\makeatother
        \begin{tabular}{cc|c}
        \toprule
        Visual & Textual          & PCK               \\
        \hline
        \checkmark &       & 91.66             \\
        \checkmark & \checkmark   &  \textbf{92.18}   \\
        \bottomrule
        \end{tabular}
\end{minipage}
    \quad
    \begin{minipage}{0.45\linewidth}
        \caption{{Ablation study of LLM model size.}}
        \vspace{-0.3cm}
        \centering
        \label{tab:modelszie}
            \setlength\tabcolsep{16pt}
            \makeatletter\def\@captype{table}\makeatother
                \begin{tabular}{c|c}
             \toprule
                LLM Size  & PCK \\
                \hline
        7B & 91.66 \\
                13B  &  \textbf{91.98} \\
       \bottomrule
                \end{tabular} 
    \end{minipage}
\end{center}
% \vspace{-0.2cm}
\end{table}

\subsection{Ablation Study}

In this subsection, we perform ablation study on the design choices of our model. 
The experiments are conducted on the visual prompt-based keypoint detection task using the MP-100 Split-1 setting with the PCK metric reported.
% The results on MP100 benchmark (Split1) are used as the performance indicator of the studied models. 

\noindent \textbf{Identify-then-Detect (ItD) Strategy.} KptLLM follows the identify-then-detect approach, where the model first interprets the keypoint's semantic information and then predicts its precise location. Tab.~\ref{tab:ablation_semantic} shows that this strategy significantly improves performance, thanks to the inter-task synergy enabled by the ItD mechanism.

\noindent \textbf{Prompt Feature Extractor.} In Tab.~\ref{tab:ablation_prompt}, we compare our prompt feature extractor (Sec.~\ref{sec:Prompt_feature}) with the average pooling based feature extraction method~\citep{xu2022pose}. The results show that our prompt feature extractor significantly outperforms the baseline method ($91.66$ vs $89.78$), which validates the efficacy of our prompt feature extractor in enhancing focus on fine-grained keypoint areas.

\noindent \textbf{Combining Visual and Textual Prompts.} In Tab.~\ref{tab:ablation_two_prompt}, rather than relying solely on the visual prompt for localization, we further incorporate the textual prompt to demonstrate the effect of this combination. The improved results indicate that the textual prompt could provide valuable high-level and semantically rich guidance, enhancing keypoint localization.

\noindent \textbf{Model Size of LLM.} By default, we use LLaVA-V1.5-7B as the base model. To explore the impact of the LLM model's size, we scale up the LLM by using LLaVA-V1.5-13B as the base model. As shown in Tab.~\ref{tab:modelszie}, increasing the model size enhances performance, demonstrating improved multimodal token processing capabilities.

\begin{figure*}[t] 
  \centering 
    \includegraphics[width=0.9\textwidth]{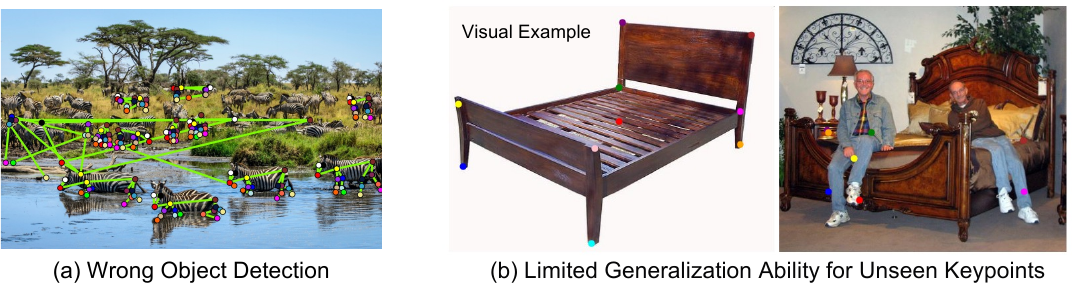}
  \caption{Failure Case Analysis.} 
  \label{fig:failure_case} 
\end{figure*}

\subsection{In-the-Wild Evaluation \& Failure Case Analysis}
\textbf{In-the-Wild Evaluation.} We demonstrate the robust keypoint comprehension capabilities of KptLLM++ by testing it on a diverse set of arbitrary images collected from the Internet. We use predefined keypoint definitions from the AP-10K dataset to prepare user instructions for querying the model. As shown in Fig.~\ref{fig:ood_test}, the visualization results highlight the strength of KptLLM++ in handling diverse image styles, out-of-distribution objects, and various complex scenes.

\noindent \textbf{Failure Case Analysis and Future Work.} We identify two main failure cases in KptLLM++: (1) \textit{Wrong Object Detection}: Accurate bounding boxes are essential for pose estimation, but even state-of-the-art detectors still make errors, such as missed objects, imprecise boxes, redundant candidates, or low confidence. These errors can severely impact keypoint detection, leading to localization failures, as shown in Fig.\ref{fig:failure_case}-(a). We expect improvements in detection models to drive breakthroughs in pose estimation. (2) \textit{Limited Generalization for Unseen Keypoints}: Despite leveraging multi-modal prompts and large-scale data, KptLLM++ struggles with generalizing to completely unseen keypoints, particularly for specific or unique objects (Fig.\ref{fig:failure_case}-(b)). This highlights the need for more diverse training data, including unsupervised methods for keypoint discovery~\citep{sun2022self,mallis2023keypoints}.

\section{Conclusion}
This paper introduces the novel problem of \textit{Generic Keypoint Comprehension}, addressing the limitations of current multimodal large language models in capturing fine-grained semantics, particularly for object keypoints. To tackle this challenge, we propose KptLLM++, a unified multimodal large language model specifically designed to process diverse modality inputs, enabling the interpretation of both semantic content and keypoint locations. By scaling the training dataset to over 500,000 samples, KptLLM++ pushes the boundaries of performance, achieving state-of-the-art results on various keypoint detection benchmarks and datasets. Through in-the-wild testing, we further demonstrate KptLLM++'s powerful generalization to complex scenarios, including diverse poses, image styles, and occlusions. This work not only provides a new solution that unifies keypoint detection across diverse contexts but also establishes a strong foundation for advancing fine-grained vision-language understanding and fostering more effective human-AI collaboration.

\section*{Acknowledgements}
The work is partially supported by the Young Scientists Fund of the National Natural Science Foundation of China under grant No.62106154, by the Natural Science Foundation of Guangdong Province, China (General Program) under grant No.2022A1515011524, and by Shenzhen Science and Technology Program JCYJ20220818103001002, and by the Guangdong Key Laboratory of Big Data Analysis and Processing, Sun Yat-sen University, China.

\section*{Declarations}

\noindent \textbf{Data Availability} All datasets used in this study are publicly available.

\noindent \textbf{Conflict of Interest} The authors declare that they have no conflicts of interest.

{\footnotesize
	\bibliographystyle{plainnat}
	\bibliography{egbib}
}
% \end{flushend}
\balance
\end{document}